\documentclass[letterpaper]{article} 
\usepackage[]{aaai24}  
\usepackage{times}  
\usepackage{helvet}  
\usepackage{courier}  
\usepackage[hyphens]{url}  
\usepackage{graphicx} 
\usepackage{natbib}  
\usepackage{caption} 
\usepackage{algorithm}
\usepackage{algorithmic}

\usepackage{newfloat}
\usepackage{listings}
\DeclareCaptionStyle{ruled}{labelfont=normalfont,labelsep=colon,strut=off} 
\floatstyle{ruled}
\newfloat{listing}{tb}{lst}{}
\floatname{listing}{Listing}
\def\paperTitle{BackTrack: Robust template update via Backward Tracking of candidate template}

\newcommand\algName{BackTrack}
\newcommand\algNameShort{BT}

\newcommand{\template}{\tb{z}} 
\newcommand{\search}{\tb{x}} 
\newcommand{\tvar}{t} 
\newcommand{\thetab}{\boldsymbol{\theta}}
\newcommand{\tracker}{\mathbf{f}_{\thetab}}
\newcommand{\trackersmall}{\mathbf{f}_{\thetab}^{s}}
\newcommand{\bbox}{\tb{b}}
\newcommand{\fwdtracklet}{\tb{B}^{Fwd}}
\newcommand{\bwdtracklet}{\tb{B}^{Bwd}}
\newcommand{\templatecandi}{\tb{z}^{*}}
\newcommand{\bwdbbox}{\bbox^{Bwd}} 
\newcommand{\tstart}{\tau_{s}}
\newcommand{\tend}{\tau_{e}}
\newcommand{\tsamprate}{k_{step}}
\newcommand{\iouzerothres}{\sigma^{thres}}

\usepackage{times}
\usepackage{microtype}
\usepackage{epsfig}
\usepackage[table,xcdraw,dvipsnames]{xcolor}
\usepackage{caption}
\usepackage{placeins}
\usepackage{color, colortbl}
\usepackage{enumitem}
\usepackage{tabularx}
\usepackage{xstring}
\usepackage{multirow}
\usepackage{xspace}
\usepackage{url}
\usepackage{subcaption}
\usepackage{xcolor}
\usepackage[hang,flushmargin]{footmisc}
\usepackage{array}
\usepackage{slashbox,pict2e}
\usepackage{soul}
\usepackage{boldline} 
\usepackage{adjustbox}

\usepackage{makecell}

\usepackage{pgfplots}
\usepackage{tikz}

\usepackage{newtxtext,newtxmath}
\newcommand{\ti}[1]{{\it{#1}}}
\newcommand{\tb}[1]{{\textbf{#1}}}

\usepackage{listings}

\definecolor{codegreen}{rgb}{0,0.6,0}
\definecolor{codegray}{rgb}{0.5,0.5,0.5}
\definecolor{codeorange}{rgb}{1,0.49,0}
\definecolor{backcolour}{rgb}{0.95,0.95,0.96}

\definecolor{color0}{rgb}{0.533333333333333,0.08,0.0823529411764706}
\definecolor{color1}{rgb}{0,1,1}
\definecolor{color2}{rgb}{1,0,1}

\definecolor{bblue}{HTML}{4F81CC}
\definecolor{rred}{HTML}{C3510A}
\definecolor{ggreen}{HTML}{9BBB90}
\definecolor{ppurple}{HTML}{9F4C7C}

\lstdefinestyle{mystyle}{
    backgroundcolor=\color{backcolour},   
    commentstyle=\color{codegray},
    keywordstyle=\color{codeorange},
    numberstyle=\tiny\color{codegray},
    stringstyle=\color{codegreen},
    basicstyle=\ttfamily\footnotesize,
    breakatwhitespace=false,         
    breaklines=true,                 
    captionpos=b,                    
    keepspaces=true,                 
    numbers=left,                    
    numbersep=5pt,                  
    showspaces=false,                
    showstringspaces=false,
    showtabs=false,                  
    tabsize=2,
    xleftmargin=10pt,
}

\definecolor{color0}{rgb}{0.533333333333333,0,0.0823529411764706}
\definecolor{color1}{rgb}{0,1,1}
\definecolor{color2}{rgb}{1,0,1}

\definecolor{colorOB}{rgb}{1,0,0} 
\definecolor{colorO}{rgb}{1,0,0} 

\definecolor{colorMB}{rgb}{0,1,0} 
\definecolor{colorMT}{rgb}{0,1,0} 
\definecolor{colorM}{rgb}{0,1,0} 

\definecolor{colorSB}{rgb}{0,0,1} 
\definecolor{colorST}{rgb}{0,0,1} 
\definecolor{colorS}{rgb}{0,0,1} 

\definecolor{colorbase}{HTML}{B3DBDB}
\definecolor{colorbase2}{HTML}{1B78AF}

\definecolor{colorbtc}{HTML}{4E7779}

\newcommand{\myUpC}[1]{{(\textcolor{colorbase2}{#1$\blacktriangle$})}}
\newcommand{\myUpR}[1]{{(\textcolor{red}{#1$\blacktriangle$})}}

\newcommand{\R}[1]{{%
    \textbf{%
        \ifstrequal{#1}{1}{\textcolor{red}{R#1}}{%
        \ifstrequal{#1}{2}{\textcolor{blue}{R#1}}{%
        \ifstrequal{#1}{3}{\textcolor{magenta}{R#1}}{%
        \ifstrequal{#1}{4}{\textcolor{teal}{R#1}}{%
                           \textcolor{cyan}{R#1}%
        }}}}%
    }%
}}

\title{\paperTitle}

\author{
    Dongwook Lee\textsuperscript{\rm 1},
    Wonjun Choi\textsuperscript{\rm 1}, Seohyung Lee\textsuperscript{\rm 1}, ByungIn Yoo\textsuperscript{\rm 1},    Eunho Yang\textsuperscript{\rm 2,3}, Seongju Hwang\textsuperscript{\rm 2,3}\\    
}
\affiliations {
    \textsuperscript{\rm 1}Samsung Advanced Institute of Technology (SAIT)\\
    \textsuperscript{\rm 2}Korea Advanced Institute of Science and Technology (KAIST), \textsuperscript{\rm 3}AITRICS\\
    \{dw23.lee, wonj4.choi, seoh.lee, byungin.yoo\}@samsung.com,\{eunhoy, sjhwang82\}@kaist.ac.kr
}

\begin{document}

\maketitle

\begin{abstract}
Variations of target appearance such as deformations, illumination variance, occlusion, etc., are the major challenges of visual object tracking that negatively impact the performance of a tracker. 
An effective method to tackle these challenges is template update, which updates the template to reflect the change of appearance in the target object during tracking.
However, with template updates, inadequate quality of new templates or inappropriate timing of updates may induce a model drift problem, which severely degrades the tracking performance. Here, we propose \algName, a robust and reliable method to quantify the confidence of the candidate template by backward tracking it on the past frames. Based on the confidence score of candidates from \algName, we can update the template with a reliable candidate at the right time while rejecting unreliable candidates. \algName~is a generic template update scheme and is applicable to any template-based trackers. 
Extensive experiments on various tracking benchmarks verify the effectiveness of \algName~over existing template update algorithms, as it achieves SOTA performance on various tracking benchmarks.
\end{abstract}

\section{Introduction}
\label{sec:intro}


Visual Object Tracking (VOT) is one of the fundamental computer vision tasks which aims to identify and locate the target object in a sequence of frames. Once the target is specified in the initial frame, the tracker finds the position of the target in the current frame, for example with a bounding box, and then finds the matching object in the consecutive frames. VOT is an essential task in various computer vision applications such as autonomous driving, video surveillance systems, and auto focusing cameras for video recording.

\begin{figure}[!h]
\captionsetup[subfigure]{labelformat=empty, justification=centering}
    \begin{subfigure}[t]{0.32\columnwidth}
    \caption{\quad STARK}
         \begin{adjustbox}{width=\columnwidth} 
\begin{tikzpicture}

\begin{axis}[
legend cell align={left},
tick align=outside,
tick pos=left,
bar width= 15pt,
xtick style={color=black},
xtick={5,10,15,20,30,50,100,200},
symbolic x coords={5,10,15,20,30,50,100,200},
xticklabels={5,10,15,20,30,50,100,200},
ylabel={AUC [\%]},
ylabel near ticks,
ymin=64, ymax=68.5,
ytick style={color=black},
legend image code/.code={
        \draw [#1] (0cm,-0.1cm) rectangle (0.2cm,0.25cm); },
]
]

\newcommand\markscale{0.8}
\newcommand\markscaletwo{1}
\newcommand\linethic{thick}


\addplot[ybar] [style={rred}]
table {%
5	64.6
10	65.5
15	65.8
};
\addplot[ybar] [style={bblue}]
table {%
20	66.2
30	66.9
50	66.2
100	66.7
200	67.0
};
\addlegendentry{STARK-ST}

\addplot [ultra thick, black, densely dotted]
table {%
5	65.8
10	65.8
15	65.8
20	65.8
30	65.8
50	65.8
100	65.8
200	65.8
};
\legend{}

\end{axis}

\end{tikzpicture}
         \end{adjustbox}
    \end{subfigure}
    \begin{subfigure}[t]{0.32\columnwidth}
         \caption{\quad MixFormer}
         \begin{adjustbox}{width=\columnwidth} 
\begin{tikzpicture}

\begin{axis}[
legend cell align={left},
tick align=outside,
tick pos=left,
bar width= 15pt,
xtick style={color=black},
xtick={5,10,15,20,30,50,100,200},
symbolic x coords={5,10,15,20,30,50,100,200},
xticklabels={5,10,15,20,30,50,100,200},
ylabel={AUC [\%]},
ylabel near ticks,
ymin=64, ymax=71,
ytick style={color=black},
legend image code/.code={
        \draw [#1] (0cm,-0.1cm) rectangle (0.2cm,0.25cm); },
]
]

\newcommand\markscale{0.8}
\newcommand\markscaletwo{1}
\newcommand\linethic{thick}


\addplot[ybar] [style={rred}]
table {%
5	64.4
10	65.6
15	65.9
20	66
30	67.5
50	67.5
};
\addplot[ybar] [style={bblue}]
table {%
100	68.1
200	69.2
};
\addlegendentry{MixFormer}

\addplot [ultra thick, black, densely dotted]
table {%
5	67.9
10	67.9
15	67.9
20	67.9
30	67.9
50	67.9
100	67.9
200	67.9
};
\legend{}

\end{axis}

\end{tikzpicture}
         \end{adjustbox}
    \end{subfigure}
    \begin{subfigure}[t]{0.32\columnwidth}
        
         \begin{adjustbox}{width=\columnwidth} 
         \hspace{1cm}
         \end{adjustbox}
    \end{subfigure}
    \hfill
    \vspace{-0.2cm}
    \hfill
    \begin{subfigure}[t]{0.32\columnwidth}
         \caption{\quad  STARK+\algNameShort}
         \begin{adjustbox}{width=\columnwidth} 
\begin{tikzpicture}

\begin{axis}[
legend cell align={left},
tick align=outside,
tick pos=left,
bar width= 15pt,
xtick style={color=black},
xtick={5,10,15,20,30,50,100,200},
symbolic x coords={5,10,15,20,30,50,100,200},
xticklabels={5,10,15,20,30,50,100,200},
ylabel={AUC [\%]},
ylabel near ticks,
ymin=64, ymax=68.5,
ytick style={color=black},
legend image code/.code={
        \draw [#1] (0cm,-0.1cm) rectangle (0.2cm,0.25cm); },
]
]

\newcommand\markscale{0.8}
\newcommand\markscaletwo{1}
\newcommand\linethic{thick}

\addplot[ybar] [style={bblue},ultra thick]
table {%
5	66.46
10	67.85
15	67.79
20	67.28
30	67.33
50	67.79
100	66.76
200	66.29
};
\addlegendentry{STARK-\algNameShort}

\addplot [ultra thick, black, densely dotted]
table {%
5	65.8
10	65.8
15	65.8
20	65.8
30	65.8
50	65.8
100	65.8
200	65.8
};
\legend{}

\end{axis}

\end{tikzpicture}
         \end{adjustbox}
    \end{subfigure}
    \begin{subfigure}[t]{0.32\columnwidth}
         \caption{\quad MixFormer+\algNameShort}
         \begin{adjustbox}{width=\columnwidth} 
\begin{tikzpicture}

\begin{axis}[
legend cell align={left},
tick align=outside,
tick pos=left,
bar width= 15pt,
xtick style={color=black},
xtick={5,10,15,20,30,50,100,200},
symbolic x coords={5,10,15,20,30,50,100,200},
xticklabels={5,10,15,20,30,50,100,200},
ylabel={AUC [\%]},
ylabel near ticks,
ymin=64, ymax=71,
ytick style={color=black},
legend image code/.code={
        \draw [#1] (0cm,-0.1cm) rectangle (0.2cm,0.25cm); },
]
]

\newcommand\markscale{0.8}
\newcommand\markscaletwo{1}
\newcommand\linethic{thick}

\addplot[ybar] [style={bblue},ultra thick]
table {%
5	70.1
10	70.2
15	70.3
20	70
30	69.9
50	69.9
100	69
200	69
};

\addplot [ultra thick, black, densely dotted]
table {%
5	67.9
10	67.9
15	67.9
20	67.9
30	67.9
50	67.9
100	67.9
200	67.9
};
\legend{}

\end{axis}

\end{tikzpicture}
         \end{adjustbox}
    \end{subfigure}
    \begin{subfigure}[t]{0.32\columnwidth}
         \caption{\quad OSTrack+\algNameShort}
         \begin{adjustbox}{width=\columnwidth} 
\begin{tikzpicture}

\begin{axis}[
legend cell align={left},
tick align=outside,
tick pos=left,
bar width= 15pt,
xtick style={color=black},
xtick={5,10,15,20,30,50,100,200},
symbolic x coords={5,10,15,20,30,50,100,200},
xticklabels={5,10,15,20,30,50,100,200},
ylabel={AUC [\%]},
ylabel near ticks,
ymin=70, ymax=73.2,
ytick style={color=black},
legend image code/.code={
        \draw [#1] (0cm,-0.1cm) rectangle (0.2cm,0.25cm); },
]
]

\newcommand\markscale{0.8}
\newcommand\markscaletwo{1}
\newcommand\linethic{thick}

\addplot[ybar] [style={bblue},ultra thick]
table {
5	72.05 
10	72.92
15	73.06
20	72.84
30	73.01
50	72.39
100	71.50
200	71.61
};

\addplot [ultra thick, black, densely dotted]
table {%
5	71.1
10	71.1
15	71.1
20	71.1
30	71.1
50	71.1
100	71.1
200	71.1
};
\legend{}

\end{axis}

\end{tikzpicture}
         \end{adjustbox}
    \end{subfigure}


  \caption{Variations of tracking performance (AUC, Area Under the Curve) according to the different template update cycles $N$ in LaSOT~\cite{fan2019lasot}. 
  Compared to the performance without template update (black dashed line), BackTrack, in short BT, consistently improves the performance of the trackers~\cite{yan2021learning, cui2022mixformer, ye2022joint} across all values of $N$ (bottom) while the previous template update methods
  show either less effective or even degradation in performance (top). }
\vspace{-0.4cm}
\label{fig:motivation}
\end{figure}
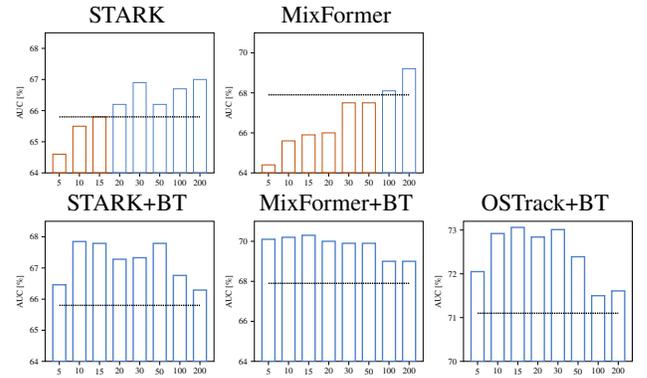

  

Earlier deep trackers~\cite{li2018high,li2019siamrpnpp,bertinetto2016fully} were mostly based on convolutional neural networks, but with the recent advent of multi-head self-attention~\cite{vaswani2017attention}, Transformer-based trackers are receiving increasing attentions due to their superior performance\cite{lin2021swintrack,zhao2021trtr,zhu2021vitt,cui2022mixformer,ye2022joint}.
Both type of trackers extract the deep features from the target object at the initial frame, which is called a \ti{template}. The features are extracted from the \ti{search} frames which we want to find the target in. Then, the features from the template and search frames are cross-correlated to find the location of the target in each frame. 
Although these architectural advancements have resulted in significant performance improvements in the object tracking task, the models are still prone to incorrect tracking due to target appearance changes, illumination variation, occlusion of the target object, and presence of similar distractors. 

\begin{figure*}[!ht]
  \centering \includegraphics[width=\textwidth]{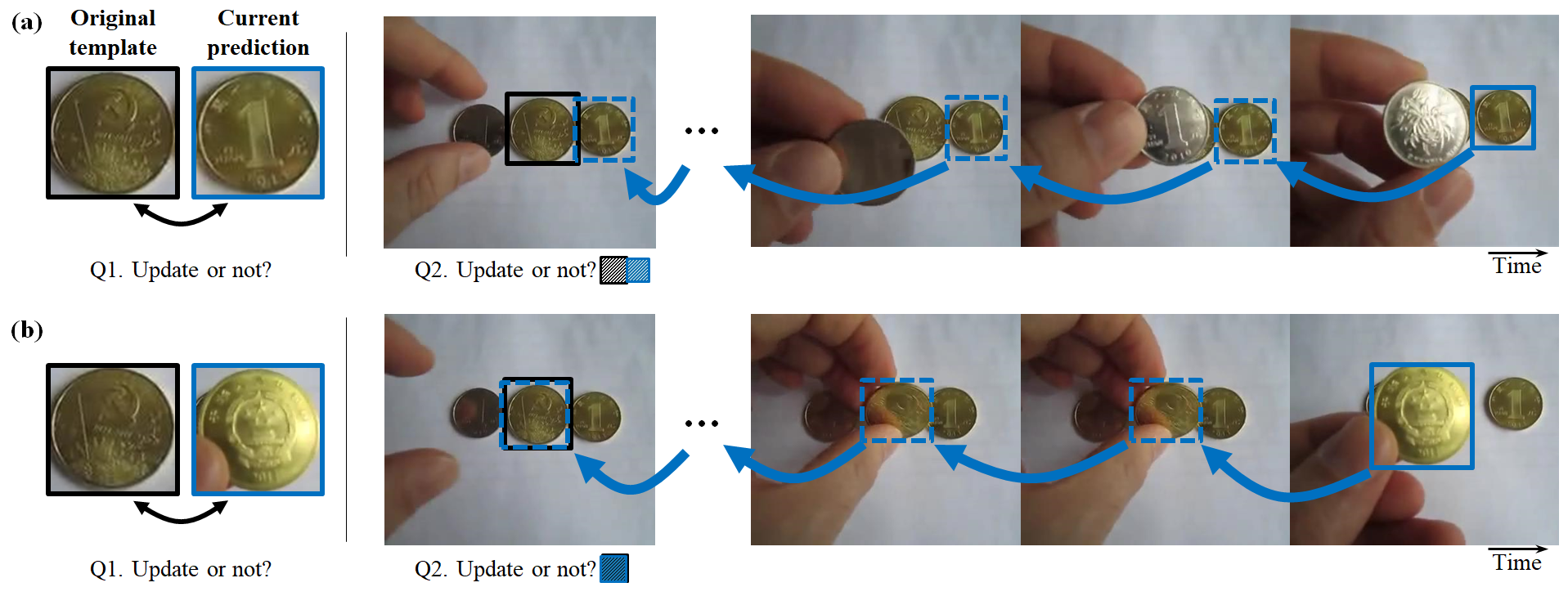}
  \vspace{-0.3cm}
  \caption{How can we decide whether to update the template or not? (a) Left: should the template be updated by the current prediction (blue box)? Most of the current template update methods might answer `Yes' because the original template (black) and the current prediction have very similar appearance. 
Right: When we track the object of current prediction in a time-reversed manner (blue arrows), the prediction at the first frame does not match the original target, and thus the template update should be rejected. 
(b) Since the resulting bounding box of backward tracking highly overlaps that of the original target, the current prediction can replace the outdated template.}
  \label{fig:concept}
  \vspace{-0.5cm}
\end{figure*}

Such temporal variances of the target make it challenging to match the target in later frames to the template, which necessitates an occasional update for the recent appearance of the target with the most recent appearance. This process is called \ti{template update}. 
Several methods~\cite{yan2021learning,liu2022tracking,cui2022mixformer, zhou2021object} have been proposed recently. 
These trackers select additional online template by learning a separate verifier, called \ti{confidence head}. The head predicts the score that is used to decide whether the current prediction of the target is appropriate as a new template or not. If the candidate template from the current prediction scores sufficiently high, the tracker replaces the outdated online template with the most recent appearance. 

%

Particularly, a template update is required more frequently and accurately as the object tracking difficulty increases. 
However, most template update models have less than perfect performance and intermittent, inaccurate template updates could lead to suboptimal templates that could severely degrade the performance such as model drift~\cite{huang2019bidirectional}. As shown in Fig.~\ref{fig:motivation} (1st row), frequent update of template update even degrades the performance due to inaccurate template updates and as a result, the performance of template update becomes highly dependent on hyperparameter tuning. 
Therefore, the existing methods update the template rarely~\cite{zhang2019learning, yan2021learning, zhang2022siamrdt} or select the best template from multiple template candidates obtained over a period~\cite{cui2022mixformer}. Although this approach may reduce inaccurate template updates, it also does not sufficiently improve the performance of the tracker.

The performance of an object tracker critically depends on the success of the template update, and the focus of this work is to come up with an effective and reliable template update strategy. How can we then determine whether a candidate is beneficial or detrimental as a new template? Existing template update methods only consider the appearance, or feature-wise similarity of the candidate template to the previous template (Figure~\ref{fig:concept}, left). However, this could fail when there exists a distractor with a large appearance similarity to the target. This is mainly because we only consider the appearance of the objects at the current video frame while disregarding their past trajectories. 

We tackle this challenge in updating the template of a tracker with an intuitive idea: backtrack the past frames with the candidate as the template by going backward in time (Figure~\ref{fig:concept}, right). 
If the candidate is a correct template, the tracklet from the backward tracking should highly overlap with its forward tracklet (Fig.~\ref{fig:concept}, bottom right), in which case we may safely update the template with it, and if not (black, Fig.~\ref{fig:concept}, top right), the candidate should be rejected.

Our \algName~(Backward Tracking of candidate template) is a generic algorithm that can enhance the performance of any base trackers by making the template update to be more reliable and robust, by preventing incorrect template updates. A potential drawback of this approach is that it requires additional computations, but we tackle this by proposing a fast, lightweight tracker since backward tracking does not have to be extremely accurate. The experimental validation verifies the effectiveness of our \algName, as it obtains state-of-the-art performance on various tracking benchmarks. Our contributions can be summarized as follows:
\begin{itemize}
    \item We propose a robust template update method, called \algName, which evaluates a template by tracking the template backward in time and comparing the resulting backtracking trajectory with the forward tracking trajectory. 
    \item Our \algName~is generally applicable to any trackers, and we further propose a lightweight backward tracker and an early termination scheme for efficient backtracking in real-time applications. 
    \item \algName~achieves state-of-the-art performance with real-time speed in various tracking benchmarks: LaSOT, TrackingNet~\cite{muller2018trackingnet}, and GOT-10k~\cite{huang2019got} in terms of robustness and more frequent updates. 
\end{itemize}

\section{Related Work}
\label{sec:related}

\subsection{Modern deep trackers} 
The Siamese network-based tracking method ~\cite{bertinetto2016fully, li2018high, zhang2022siamrdt} shows a significant improvement in performance compared to the previously used correlation filter-based tracker. However, it localizes a target by relying on correlation based on simple operations. Thus, it is vulnerable to distractors and is disadvantageous. Recently, attention-based transformers~\cite{vaswani2017attention, dosovitskiy2020image, liu2021swin} have been adopted for vision tasks. Consequently, many tracking methods~\cite{zhao2021trtr, yan2021learning, chen2021transformer} have replaced the correlation to transformers. In addition, SwinTrack~\cite{lin2021swintrack} adopted the Swin-Transformer~\cite{liu2021swin} as a backbone instead of CNN and Transformer~\cite{vaswani2017attention}. Transformer is effective for feature fusion as it is capable of long-term dependency. Additionally, it is more robust in representation learning than the CNN-based backbone and significantly improves the performance.

Recently, the multi-staged pipeline of conventional trackers was simplified by unifying the process of feature extraction and target information integration. With a unified framework, target-specific feature extraction is possible and improved correlation can be expected. MixFormer~\cite{cui2022mixformer} proposed a Mixed Attention Module for simultaneous feature extraction and target information integration. OSTrack~\cite{ye2022joint} proposed the one-stream method by concatenating the patch queries from the template the search. The feature of the search frame is extracted adaptively to the template. 


\subsection{Template update} 
%
STARK-ST~\cite{yan2021learning} utilizes an additional confidence head based on the similarity between the templates and new candidate template. Likewise, MixFormer~\cite{cui2022mixformer} yields a similarity score of the candidate template from each frame in a given interval and choose the one with best score as a new online template.
Specifically, STARK-ST outputs the confidence score every $N$-th frame and decides whether to update the template or not by comparing the score with a given threshold. Contrarily, MixFormer calculates the confidence score for every frame and stores the pairs of candidates and their corresponding confidence score. For every $N$-th frame, among the $N$ pairs of the templates and scores, the candidate template with the best confidence score is chosen to as the new template. Consequently, hyper-parameter $N$ which represents the period of template update affects performance in both trackers.
Moreover, these types of algorithms still suffer from discriminating unsuitable templates such as distractors as they are based on the frame-by-frame feature comparison.
Further, the template will not be updated if it is drastically different from the original template, even though the candidate is the right prediction of the target object.

\begin{figure}[h!]
  \centering \includegraphics[width=0.95\columnwidth]{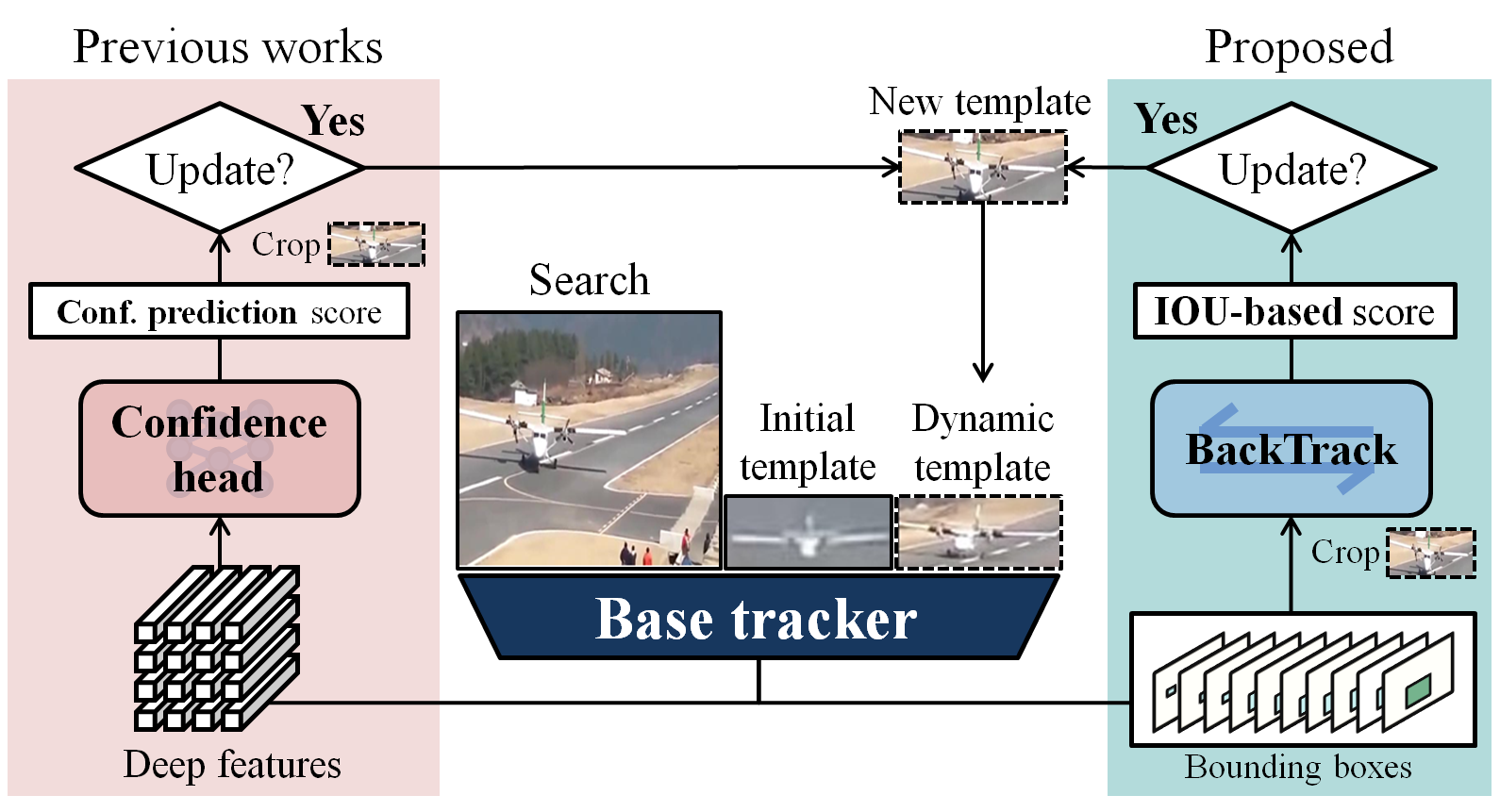}
  \vspace{-0.3cm}
  \caption{Comparison of template update mechanisms presented in previous works (left) and \algName~(right). The previous models predict the score according to the deep feature input by the confidence head, which was trained for binary classification. The proposed algorithm quantifies the intersection over union (IOU)-based score using the bounding boxes to decide whether the candidate template should be accepted or rejected.}
  \label{fig:comp_prev}
\end{figure}

Some other types of template update use Reinforcement Learning (RL), Long Short-Term Memory (LSTM), etc.~\cite{sun2020fast,yang2018learning, zhang2022siamrdt}, to choose reliable templates.
These methods require additional modules including learnable parameters for template updates. In addition, as the template update is decided by the module, the update quality depends on the module's performance which shares the limitation with other template update methods~\cite{yan2021learning,cui2022mixformer}. As shown in Fig.~\ref{fig:comp_prev} left, these kinds of method including STARK \& MixFormer predict the score of similarity between the templates and a candidate from the search based on the extracted deep features. However, the proposed method make a decision of template update quantitatively based on the comparison of forward \& backward tracklets with IOU score (Fig.~\ref{fig:comp_prev} right). 

\begin{figure*}[!ht]
  \centering \includegraphics[width=0.97\textwidth]{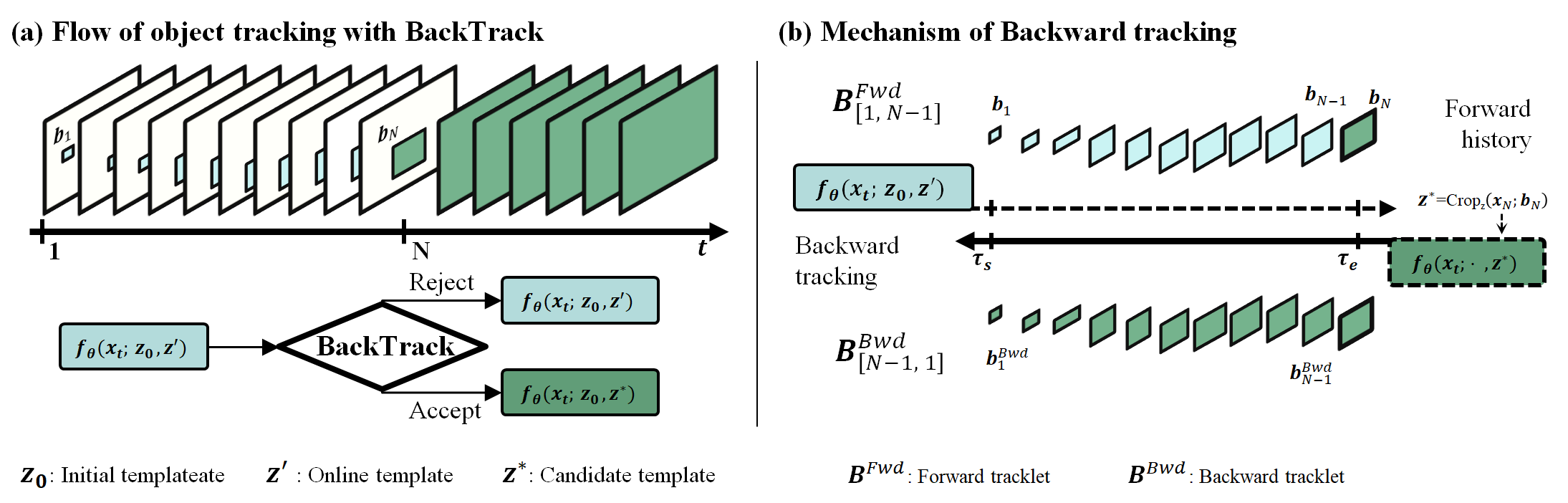}
  \vspace{-0.3cm}
  \caption{Flow and mechanism of Backward tracking}
  \label{fig:flow_BTC}
  \vspace{-0.5cm}
\end{figure*}

\subsection{Revisiting backward track} 
Tracking backward, a pivotal concept in numerous visual tracking applications, has demonstrated its utility in diverse type of applications including tracker optimization as a regularization constraints~\cite{wu2007robust}, the detection of tracking failures~\cite{kalal2010forward}, and self-supervised/unsupervised learning methodologies for pseudo labeling or consistency loss for training the deep models~\cite{wang2019unsupervised,yuanw2020self, yuand2020self, bastani2021self,wang2019learning}.
Here, to enforce the concept of cycle-consistency for self-supervised/unsupervised manner, video segmentation/tracking proceed on unlabeled frames and come back to the labelled starting frame 
to calculate the loss of the segmentation/tracking results.


Compared to these studies, the proposed method is the first attempt to apply the backward tracking concept for template update for template-based visual object tracker at the test time. The concept of cycle-consistency was only used in the training phase. By contrast, the proposed method uses the backward tracking only in the test phase. The cycle-consistency loss only focuses on how to generate a pseudo-label to compute the proper loss function without sufficient ground-truth labels. Therefore, their backward tracking is no longer useful in the test phase. On the other hand, the proposed method does not use backward tracking in the training phase. \algName~focuses on the strategy of online template update for a robust visual object tracking. This exploiting cycle-consistency at test time to check for the reliability of the candidate template for template-based visual object tracker is highly novel idea and the proposed method is considerably different from the previous methods with a backward tracking~\cite{yuanw2020self, yuand2020self, bastani2021self,wang2019learning}. 

\section{Proposed Method}

In this section, we propose our novel online template update strategy, \algName. The underlying idea is to evaluate each candidate template by backtracking it and then comparing its trajectory with the forward tracking trajectory. First, we explain the forward and backward tracking procedures. Then, we describe the proposed quantification method for measuring how good the backward tracklet is. Lastly, we describe how to enhance the efficiency of the backward tracker for real-time applications.
\label{sec:method}



\subsection{Forward tracking}
As shown in Fig.~\ref{fig:flow_BTC} top left, the tracker $\tracker$ predicts the position of the target object in the search frame using the given templates; one is the fixed template $\template_0$ from the initial frame and the other is the online template $\template'$ which is updated during inference. The tracker $\tracker$ can be written as conditional function of $\template_0$ and $\template'$ as follows,
\begin{equation}
 \bbox_{\tvar} = \tracker(\search_{\tvar};  \template_0, \template'),
\end{equation}
where $\bbox_{\tvar}$ is the output bounding box for $\tvar$-th frame and $\search_{\tvar}$ is the $\tvar$-th search frame input. The online template $\template'$ is initialized as $\template_{0}$ before any updates. The search window $\search_\tvar$ is defined by cropping the pre-defined size of ROI using the center of the previous bounding box prediction. 
For notational simplicity, we omit these causality conditions in the later equations.

The tracker proceeds for $N$ frames, obtaining the forward tracklet $\fwdtracklet_{[1,N]}$ which is a set of forward bounding boxes,
\begin{equation*}
 \fwdtracklet_{[1,N]} = [\bbox_{1}, ... , \bbox_{N}] =  [\tracker( \search_{1}; \template_{0}, \template' ), ... , \tracker( \search_{N}; \template_{0}, \template' )].
\end{equation*}
where $N$, the number of frames for the cycle of template update, is a hyperparameter. After tracking $N$ frames, we obtain a new candidate template $\templatecandi_{N}$ using the $\tvar$-th bounding box prediction $\bbox_{N}$. We then decide whether to update the template or not using \algName~, which is described in the following section.

\subsection{Backward tracking}
\algName~evaluates the quality of $\templatecandi_{N}$ as an online template by tracking backward using the candidate template $\templatecandi_{N}$ as the initial template replacing the original template $\template_{0}$ as follows:
\begin{equation}
 \bwdbbox_{\tvar} = \tracker(\search_{\tvar}; \cdot, \templatecandi_{N}),
\end{equation}
where $\bwdbbox_{\tvar}$ is the output bounding box for $\tvar$-th frame in backward tracking.
The backward tracker goes back in time by going through the past frames in a reverse order. It yields a backward tracklet $\bwdtracklet_{[N-1,1]}$ (Fig.~\ref{fig:flow_BTC} bottom right):

\begin{align*}
 \bwdtracklet_{[N-1,1]} &=  [\bwdbbox_{N-1}, ... , \bwdbbox_{1}] \\
   &=  [\tracker(\search_{N-1}; \cdot, \templatecandi_{N}), ... , \tracker(\search_{1}; \cdot, \templatecandi_{N}) ].
\end{align*}
Note that $\bwdtracklet$ starts from $N-1$ because $\templatecandi_N$ was extracted from $\search_N$ and the tracker $\tracker$ shares the weight parameters $\thetab$ between the forward pass and the backward pass. 
\algName~is performed at every $N$-th frame during the forward tracking.

\subsection{Quantification of confidence score} 
The template update of previous works~\cite{yan2021learning,cui2022mixformer} are based on the confidence head which simply compares the similarity between the template and candidates (Fig.~\ref{fig:comp_prev}). However, we 
 propose a more trustworthy scoring scheme based on Intersection Over Union (IOU) between two tracklets, where one is a forward tracklet $\fwdtracklet_{[1,N-1]}$ and the other is a backward tracklet $\bwdtracklet_{[N-1,1]}$. 
There are several options in how to use the IOU between $\fwdtracklet_{[1,N-1]}$ and $\bwdtracklet_{[N-1,1]}$ for scoring. We empirically found that the combination of the following two metrics achieve the best performance. 

The first metric is the count of bounding boxes which have sufficiently large IOU between $\fwdtracklet_{[1,N-1]}$ and $\bwdtracklet_{[N-1,1]}$
\begin{equation}
    \sum_{\tvar=1..N}[IOU(\fwdtracklet_\tvar,\bwdtracklet_\tvar)>0.5] 
    \label{eq:cond2}
\end{equation}
where $[\cdot]$ is an indicator function. We consider the case where $IOU > 0.5$ as a hit 
which we assume that the backward tracker follows the forward tracklets in the reverse order well. 

The second metric is the IOU between the forward prediction $\bbox_1$ and the last backward prediction $\bwdbbox_1$ (here, $\tstart$=1), 
\begin{equation}
    IOU(\fwdtracklet_{\tstart} ,\bwdtracklet_{\tstart}). 
    \label{eq:cond1}
\end{equation}
We set the threshold $\iouzerothres$, as high as 0.9. 
and it requires a new template to be more precise and reliable. 


\begin{table*}[!ht]
\centering
\caption{Comparison state-of-the-art results on various object tracking benchmarks. The numbers between parenthesis represent the performance improvement compared to each base tracker without an online template update. (blue: a template update method, red: the proposed method \algName, in short \algNameShort). 
All the performances are reproduced results based on the code and the weights from each GitHub link. }
\begin{adjustbox}{width=\linewidth,center}
\begin{tabular}{l|c|ll|ll|ll|ll|lll} 
\hlineB{3}
\multirow{2}{*}{Method}  & \multirow{2}{*}{\thead{Template \\ update}} & \multicolumn{2}{c}{LaSOT} & \multicolumn{2}{c}{LaSOT$_{ext}$} & \multicolumn{2}{c}{TrackingNet}& \multicolumn{2}{c}{OTB100} & \multicolumn{3}{c}{GOT-10k} \\ 
\cline{3-13}
 & & AUC & $P_{Norm}$ & AUC & $P_{Norm}$ &AUC & $P_{Norm}$ & AUC & $P_{Norm}$ & AO & SR$_{0.5}$ &SR$_{0.75}$ \\ 
\hline
STARK-S &  No & 65.8 & 75.2 &  46.7 & 52.8 & 80.3 & 85.1  & 65.9          & 80.9 & 67.2 &76.1 &61.2\\
STARK-ST & Yes & 66.4\myUpC{0.6} & 76.3\myUpC{1.1}& 47.4\myUpC{0.7} & 53.9\myUpC{1.1} &
81.3\myUpC{1.0} & 86.1\myUpC{1.0}& 67.6\myUpC{1.7} & 83.5\myUpC{2.6} & 68.0\myUpC{0.8} & 77.7\myUpC{1.6} & 62.3\myUpC{1.1} \\
STARK-\algNameShort & Proposed & 67.8\myUpR{2.0} & 78.1\myUpR{2.9} &  49.0\myUpR{2.3} & 55.8\myUpR{3.0} &
81.4\myUpR{1.1} & 86.3\myUpR{1.2}& 69.1\myUpR{3.2} & 84.7\myUpR{3.8} & 69.8\myUpR{2.6} & 80.0\myUpR{3.9} & 64.5\myUpR{3.2} \\
\cline{1-13}
MixFormer &  No & 67.9 & 77.6 &  50.9 & 61.6 & 80.8 & 85.6  & 67.6          & 82.2 & 69.8 & 78.9 & 65.7 \\
MixFormer &  Yes & 69.2\myUpC{1.3} & 78.7\myUpC{1.1} &  51.0\myUpC{0.1} & 61.7\myUpC{0.1} &
83.1\myUpC{2.3} & 88.1\myUpC{2.5}  & 70.1 \myUpC{2.5} & 86.1\myUpC{3.9}& 70.7\myUpC{0.9} & 80.0\myUpC{1.1} & 67.8\myUpC{2.1} \\
MixFormer-\algNameShort & Proposed & 70.3\myUpR{2.4} & 80.5\myUpR{2.9} & 52.5\myUpR{1.6} & \tb{63.5}\myUpR{1.9} &
82.8\myUpR{2.0} & 87.9\myUpR{2.3} & 70.3 \myUpR{2.7} & 85.5\myUpR{3.3}& 74.8\myUpR{5.0} & 85.0\myUpR{6.1} & 72.6\myUpR{6.9} \\
\cline{1-13}
OSTrack &  No & 71.1 & 81.1 & 50.9 & 58.1 &83.9 &88.5   & 68.7             & 83.6& 73.7 & 83.2 & 70.8 \\ 
OSTrack-\algNameShort  & Proposed & \tb{73.3}\myUpR{2.2} & \tb{83.7}\myUpR{2.6}&  \tb{53.3}\myUpR{2.4} & 60.6\myUpR{2.5} &
\tb{85.0}\myUpR{1.1} & \tb{89.7}\myUpR{1.2}  & \tb{70.4}\myUpR{1.7} & \tb{86.3}\myUpR{2.7} & \tb{75.9}\myUpR{2.2} & \tb{86.9}\myUpR{3.7} & \tb{73.3}\myUpR{2.5}\\
\cline{1-13}
\hlineB{3}
\end{tabular}
\end{adjustbox}
\label{table:SOTA}
\end{table*}

\subsection{Efficient backward tracking}
\subsubsection{Range \& step-size of backward track }
Every $N$-th frames, our algorithm checks whether to update the template or not.
If \algName~decides to accept the new template for the range [$\tstart$, $\tend$], the range for the next \algName~will be from $\tend$ to $\tend+N$. 
Therefore, the previous frames before the current $\tvar=\tend$ need not to be considered for computational simplicity. 

On the other hand, when \algName~rejects the candidates template for the range [$\tstart$, $\tend$], the range for the next \algName~cycle will be from $\tstart$ to $\tend+N$ which has the length of $2N$. To maintain the number of frames tracked in backward regardless of the number of rejections, the step-size $\tsamprate$ should be increased (e.g.$\tsamprate=2$ for $2N$) when the rejection occurs. If we set the range and the step-size of the \algName~as [$\tstart$:$\tstart+N\tsamprate:\tsamprate$], then the number of frames tracked backward in \algName~is kept as $N$. As a result, the inference time for \algName~remains constant.
\subsubsection{Early rejection \& termination} 
If the candidate has the resolution less than the input size of the template, \algName~rejects it without any computation of backward tracking (refer to the supplementary material for the qualitative reasoning) because the candidate template is already blurred. The backward tracking stops whenever the backward tracklet does not follow the forward tracklet. Further tracking backward until it reaches the first frame of the forward tracklet is not necessary.

\subsubsection{Backward track with small tracker}
Here, we demonstrate \algName~with the small tracker with much lighter computations. 
The purpose of the backward track is the confirmation of whether template-update is beneficial or not. Thus, it is not necessary to find the accurate bounding box of the target. It is sufficient to locate the target vaguely by distinguishing the distractors. For this purpose, the computationally smaller tracker $\trackersmall$ is utilized for backward tracking, for improved computational efficiency for real-time tracking.


\section{Experiments}
\label{sec:results}
\subsection{Experiment details} 
The proposed model is evaluated on the following VOT benchmarks: LaSOT, LaSOT$_{ext}$~\cite{fan2021lasot}, TrackingNet, OTB-100~\cite{Yi2015object}, and GOT-10k. 
The experiments are run on the following platform with Intel(R) Xeon(R) Gold 6142 CPU and NVIDIA A100 GPU.

\subsection{Base trackers}
\algName~is a generic algorithm to boost up the performance of base tracker. To investigate the effects of the proposed algorithm, we applied \algName~to the recently proposed deep trackers; STARK (STARK-R50), MixFormer (MixFormer-Base-22k), and OSTrack (OSTrack-384).
The trackers are implemented in Python using PyTorch. 

\subsection{Template updates}
STARK and MixFormer have their own template update mechanism using the confidence head. For these deep trackers, we set the base models as ones without the online template update module; STARK-S and MixFormer (w/o template update). Then we applied \algName~to the base models: STARK-\algNameShort~and MixFormer-\algNameShort. To verify the effectiveness of our \algName~, we compared the performances of the base models with \algName~with the performance of the model with their own template update modules; STARK-ST and the original MixFormer. These trackers checks for template updates at every 200-th frames ($N$=200), whereas \algName~works more frequently ($N$=10 or 15). As these trackers have dual template option, \algName~do not need any additional weights or fine-tuning process 

On the other hand, OSTrack did not utilize the dual template for the template update. Therefore, we redesigned this tracker with a simple modification of introducing another online template with its own positional encoding. We then fine-tuned the tracker with the pre-trained weights for 20 epochs with the identical training procedure of each trackers~\cite{lin2021swintrack,ye2022joint}. This modification allows us to use \algName~to update the online template of OSTrack-\algNameShort.

\begin{figure*}[!ht]
  \centering
  \includegraphics[width=0.93\linewidth]{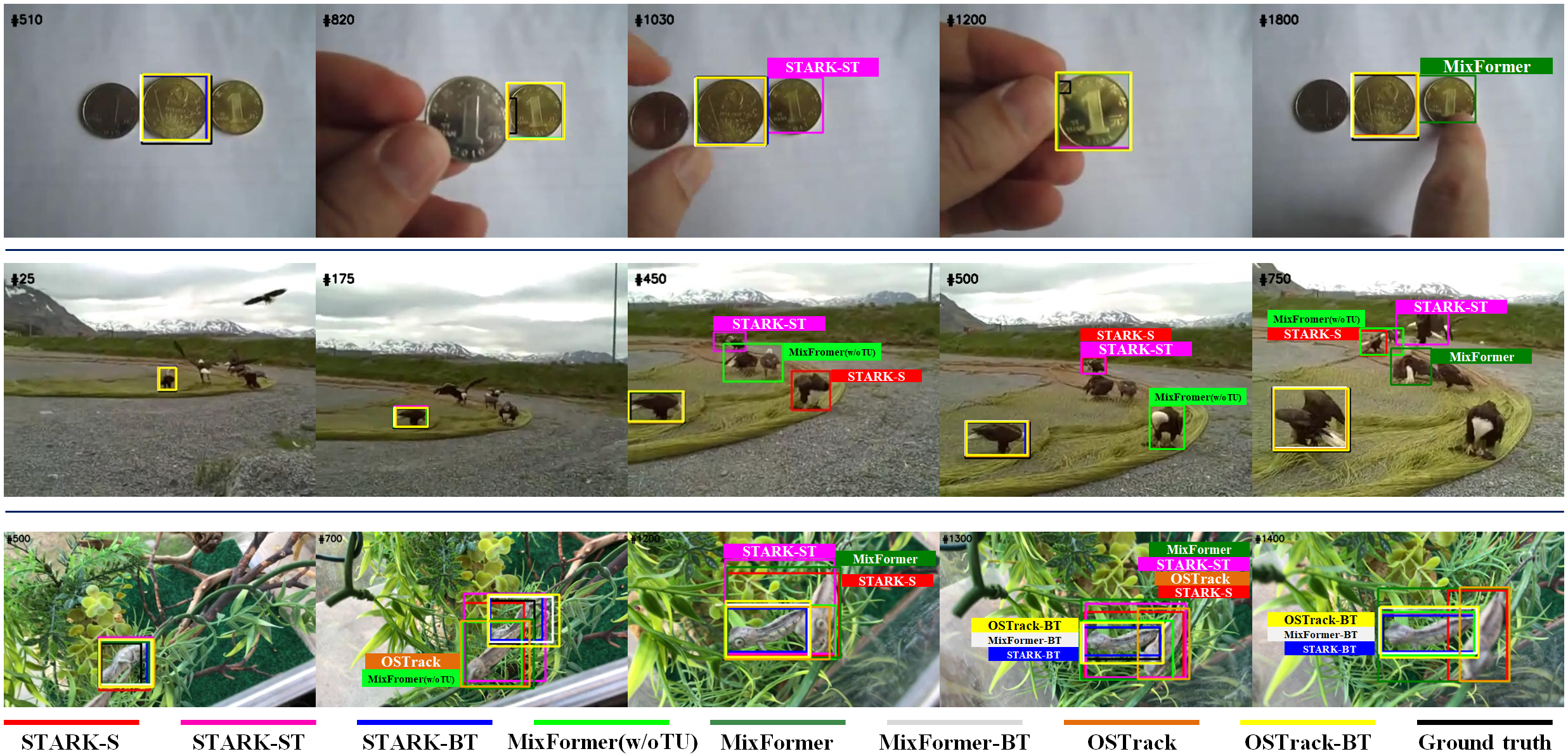}
  \vspace{-0.3cm}
  \centering
  \caption{Qualitative comparison between the proposed method and the other SOTA trackers at LaSOT (Top:`Coin-6', Middle:`Bird-15', and Bottom:`Chameleon-20'). Other template update algorithms fails to distinguish between target\&distractors or to draw accurate bounding boxes because their update module cannot reject the unsuitable template (e.g. distractor and/or occluded target) while \algName~helps each tracker updates appropriate template and timing. 
  }
  \vspace{-0.3cm}
  \label{fig:qres}
\end{figure*}


\subsection{Performance gain by \algName}
Table.~\ref{table:SOTA} shows the performance of the various SOTA trackers on LaSOT, LaSOT$_{ext}$, TrackingNet, OTB-100, and GOT-10k. As shown in the results of LaSOT, the proposed method (+\algNameShort) improved all performances of the base trackers by approximately 2.0-2.6\% in AUC and 2.6-2.9\% in $P_{Norm}$ (normalized precision). 
Compared to STARK-ST \& MixFormer which utilize the confidence head for the period of $N$ frames, the proposed backward tracking improved the performance with significantly larger gain compared to the original template update methods. 
We observe similar performance gains with LaSOT, TrackingNet, OTB-100 and GOT-10k. Although the performance improvement from the proposed algorithm with MixFormer is relatively less in TrackingNet, \algName~shows robust and consistent improvements. The superior performance improvement caused by \algName~is shown through not only the AUC \& $P_{Norm}$ but also the overlap precision \& distance precision plots in the Supplementary material.
Applying our template update method to OSTrack384 achieved 73.3 AUC and 83.7 normalized precision which are the state-of-the-art scores on LaSOT. Specifically, with combination of \algName~ and the base trackers, we achieved state-of-the-art performance on three tracking benchmarks; LaSOT (OSTrack-\algNameShort~AUC 73.3\%), TrackingNet (OSTrack-\algNameShort~AUC 85.0\%), and GOT-10k (OSTrack-\algNameShort~AO 75.9\%, one-shot setting).

\subsection{\algName~on GOT-10k} 
The evaluation on GOT-10k is done using a unique protocol called \emph{One-shot protocol}~\cite{huang2019got} which differs from the evaluation protocol on other tracking datasets. While LaSOT and TrackingNet have completely overlapping set of classes between training and test set, in GOT-10k, there is no overlap between the set of classes in the training and test sets. This is done to avoid evaluation bias towards familiar object classes to put more emphasis on the generalization performance of the tracker~\cite{huang2019got}. 

STARK-ST and MixFormer with their confidence head achieve less than 1.0\% AO improvements (Table.~\ref{table:SOTA} GOT-10k), on the targets from the unseen classes with the \emph{One-shot protocol}. Their confidence head outputs the score for the update by comparing the initial template $\template_0$ and candidate templates $\templatecandi$ which is not in the classes that are seen by the network. It is much more difficult to output reliable confidence scores on them. On the other hand, the proposed method calculates the confidence score based on the IOU metric between forward and backward tracklet which is quantitative and less dependent on the class of the target, thus is less affected by whether the target is from the seen classes or not. Therefore, \algName~shows significantly larger improvements when evaluating with One-shot protocol. 

\subsection{Representative results}
The qualitative examples for comparison with baseline template update methods are presented in Fig.~\ref{fig:qres}.
In Fig.~\ref{fig:qres} (top), the target underwent the hard occlusion, and the template update should have been rejected during that time. After the occlusion, STARK-ST(pink) and MixFormer(dark green) failed to recover to the target because the template was replaced by the distractor on the right of the target during the occlusion. These types of false positive updates occurred often with the previous approaches because the target and distractor appear similar. In Fig.~\ref{fig:qres} (middle, `Bird-15'), STARK-ST(pink) and MixFormer(dark green) confuse the target and distractors because the appearance of the target changes too fast. By contrast, STARK-\algNameShort(blue) and MixFormer-\algNameShort(grey) stuck to the target since more frequent template update also improves the ability to distinguish between the target and distractors, especially with such a rapid change in the target appearance. Additionally, more frequent update of the template draws more accurate bounding box predictions when the distractors occlude the target (Fig.~\ref{fig:qres}, bottom, `Chameleon-20').

\subsection{Efficient backward tracking} 

\subsubsection{Early rejection \& termination} Without early rejection or early termination, OSTrack384-\algNameShort runs at 29.0 FPS, which is less than real-time speed, 30 FPS. As shown in Table.~\ref{table:ablation_EE}, early rejection reduces 14.0\% of computations from the backward track with no performance degeneration, which enhances the speed to 35.3 FPS. 
Because the low resolution template is the low quality candidate, it will be certainly rejected at the end, it does not affect the overall performance.
The early termination also saves 27.9 \% of the computation from \algName~after the early rejection. Therefore, using both techniques, we reduces 38.0\% of the overall computations, and the speed is accelerated from 29.0 FPS to 39.2 FPS.
\begin{table}[h!]
\caption{Ablation study of early rejection and early termination for \algName. The experiments are performed on LaSOT with OSTrack384. 
The number between parenthesis represent the improvement of FPS compared to BackTrack w/o early rejection \& termination.
}
\vspace{-0.3cm}
\begin{adjustbox}{width=\linewidth,center}
\begin{tabular}{c|cc|lll} 
\hline
Model & Early rejection &  Early termination  &  FPS & AUC \\ 
\hline
OSTrack &            &            &  64.9 & 71.1 \\ 
OSTrack-BT &            &            &  29.0 & 73.3 \\ 
OSTrack-BT & \checkmark &            &  35.3\myUpR{6.3} & 73.3 \\ 
OSTrack-BT & \checkmark & \checkmark & 39.2\myUpR{9.8} & 73.3 \\ 
\hline
\end{tabular}

\end{adjustbox}
\label{table:ablation_EE}
\end{table}

\subsubsection{\algName~by smaller tracker} 
For backward tracking with \algName, a high-performance tracker is not necessary (Table.~\ref{table:lighter}), since a small and cost-efficient tracker can decide whether the given template is good or bad, as long as it can distinguish the distractor from the target. 
The \algName64 has only 3\% computational overhead on top of the computational cost for the base tracker, and thus we can still perform real-time tracking with OST384-\algNameShort64 (36.2 FPS). Here, the early termination and the early rejection are not applied.
The inference speed is tested including the time for data loading and pre-processing. 

\begin{table}[!h]
\centering
\caption{Comparison results of efficient \algName~models on LaSOT. The baseline model is OSTrack384 (Forward). Its variants with different input resolution are utilized as the backward tracking models (Backward). MACs show the computational cost per frame of the trackers. 
}
\vspace{-0.2cm}
\begin{adjustbox}{width=\linewidth,center}
\begin{tabular}{c|c|ll|rr} 
\hline
Forward & Backward & AUC  & $P_{Norm}$ & MACs &  FPS\\ 
\hline
OST384 & -- & 71.1 & 81.1       & 60.6 G       & 64.9 \\
OST384   & OST64  & 72.9\myUpR{1.8} & 83.2\myUpR{2.1} & 62.5 G  & 36.2\\
OST384   & OST128  & 73.3\myUpR{2.2} & 83.5\myUpR{2.4} & 67.4 G & 35.4 \\
OST384   & OST384  & 73.3\myUpR{2.2} & 83.7\myUpR{2.6} & 98.2 G & 29.0 \\
\hline
\end{tabular}

\end{adjustbox}
\label{table:lighter}
\end{table}

\subsection{Effect of template update frequency}


As shown in Fig.~\ref{fig:motivation}, the proposed method shows a stable improvement in performance regardless of $N$ (the cycle of template update) while more frequent updates of template by the previous works \cite{yan2021learning,cui2022mixformer} reduced the performance of the tracker. STARK and MixFormer suffer from a large number of false positive template updates. Here, false positive of the template update means that the confidence module allows to update with a suboptimal template which degrades the performance of the tracker in the subsequent frames. Therefore, the template update should not be used more frequently. Due to such lack of robustness, STARK and MixFormer update the template with a long period (e.g. $N=200$ frames). Further, their performance varies significantly for the hyperparameter $N$ and sometimes show performance degradation with a suboptimal $N$. However, \algName~shows consistent improvements in AUC over the base tracker regardless of $N$. These results strongly support the robustness of \algName~which is a crucial factor for the template update method.

\subsection{Ablation studies for update condition}
Table.~\ref{table:ablation} shows the results of the ablation study for the template update. We chose the combination of the IOU of the $\tstart$-th frame and the number of frames which has larger IOU than the threshold by experiments because these two metrics show complementary results. 
\begin{table}[h!]
\caption{Ablation study of template update condition for \algName. 
}
\vspace{-0.3cm}
\begin{adjustbox}{width=0.95\linewidth,center}
\begin{tabular}{c|c|ll} 
\hline
Model &Update condition &  AUC  & $P_{Norm}$  \\ 
\hline
OSTrack & Baseline & 71.1 & 81.1\\
OSTrack-BT & Eq.~\ref{eq:cond2}  & 72.0\myUpR{0.9} & 81.9\myUpR{0.8}  \\
OSTrack-BT & Eq.~\ref{eq:cond1} & 72.8\myUpR{1.7} & 82.7\myUpR{1.6}  \\
OSTrack-BT & Eq.~\ref{eq:cond2} \& Eq.~\ref{eq:cond1} & \tb{73.3}\myUpR{2.2} & \tb{83.7}\myUpR{2.6}  \\
\hline
\end{tabular}

\end{adjustbox}
\label{table:ablation}
\end{table}

In Table.~\ref{table:ablation}, condition A measures the robustness of tracker. If the backward and forward tracklet have larger discrepancy than M$^{thres}$, then template update is rejected by condition A. And condition B determines whether the backward tracklet accurately matches the target object. With combination of the condition A and B, \algName~achieves significant improvement, since it reduces the update of false positive templates, and the template is updated when the backward tracklet and the forward tracklet exactly match.

\subsection{Evaluation on VOT2022 challenge}
\algName~scores on par with the best-ranking in \emph{Unsupervised} reference scores at the short-term bounding box regression track of VOT2022 challenge~\cite{Kristan2022vot}.
\emph{Unsupervised} scores is a usual evaluation setting of the VOT benchmarks which is
an evaluation without a \ti{short-term failure recovery}. \ti{Short-term failure recovery} uses the ground truth information in the test when it fails several frames. 
For more details about the architecture of the tracker, the challenge results and the concept of \ti{short-term failure recovery}, please refer to the supplementary material.

\section{Conclusion}
\label{sec:conclusion}
We proposed a novel backtracking method to enhance the robustness of template updates in a tracker. Specifically, for each candidate template, we use it as an initial template and track the input frames backward, and compute the similarity between the forward and backward tracklets to determine whether it could serve as a reliable template for tracking. The proposed method is generally applicable to enhance the performance of any off-the-shelf deep trackers as an add-on module, with negligible computational overhead over the original tracker due to the use of a lightweight backward tracker. We showed that combining our method with existing trackers yields state-of-the-art performance on multiple tracking benchmarks, and that it allows to robustly track targets at the presence of difficult occluders.


\bibliography{9224.bbl}
\clearpage
\appendix
\label{sec:appendix}
\section{Supplementary materials}
\subsection{Success and normalized precision on LaSOT}
\begin{figure}[!ht]
   \centering
   \resizebox{0.85\linewidth}{!}{
\begin{tikzpicture}

\definecolor{color0}{rgb}{0.533333333333333,0,0.0823529411764706}
\definecolor{color1}{rgb}{0,1,1}
\definecolor{color2}{rgb}{1,0,1}

\begin{axis}[
legend cell align={left},
legend style={fill opacity=1.0, draw opacity=1,text height=7, text opacity=1, at={(0.03,0.03)}, anchor=south west},
tick align=outside,
tick pos=left,
title={\(\displaystyle \bf{Success}\)},
x grid style={white!69.0196078431373!black},
xlabel={Overlap threshold},
xmajorgrids,
xmin=0, xmax=1,
xtick style={color=black},
xtick={0,0.2,0.4,0.6,0.8,1},
xticklabels={0.0,0.2,0.4,0.6,0.8,1.0},
y grid style={white!69.0196078431373!black},
ylabel={Overlap Precision [\%]},
ylabel near ticks,
ymajorgrids,
ymin=0, ymax=91,
ytick style={color=black},
every axis plot/.append style={ultra thick, line width=1.5pt},
]
\addplot [ colorOB]
table {%
0 90.7357940673828
0.0500000007450581 90.1453628540039
0.100000001490116 89.6676330566406
0.150000005960464 89.2907791137695
0.200000002980232 88.9419860839844
0.25 88.5378799438477
0.300000011920929 88.0759506225586
0.349999994039536 87.6105041503906
0.400000005960464 87.0621643066406
0.449999988079071 86.3764953613281
0.5 85.5496139526367
0.550000011920929 84.4577331542969
0.600000023841858 82.9983062744141
0.649999976158142 80.7423400878906
0.699999988079071 77.4561386108398
0.75 72.85546875
0.800000011920929 65.8904724121094
0.850000023841858 54.9161834716797
0.899999976158142 37.1160125732422
0.949999988079071 10.9784717559814
1 0
};
\addlegendentry{\textbf{OSTrack-\algNameShort [73.3]}}
\addplot [ colorO, densely dotted]
table {%
0 88.9744491577148
0.0500000007450581 88.060905456543
0.100000001490116 87.4368133544922
0.150000005960464 86.9743499755859
0.200000002980232 86.5433044433594
0.25 86.0639419555664
0.300000011920929 85.4605941772461
0.349999994039536 84.9374465942383
0.400000005960464 84.2694625854492
0.449999988079071 83.5148086547852
0.5 82.5479049682617
0.550000011920929 81.4104766845703
0.600000023841858 79.9083480834961
0.649999976158142 77.662712097168
0.699999988079071 74.7086791992188
0.75 70.4275512695312
0.800000011920929 63.5809135437012
0.850000023841858 52.7781677246094
0.899999976158142 35.9425735473633
0.949999988079071 11.3293838500977
1 0
};
\addlegendentry{OSTrack [71.1]}
\addplot [ colorMB]
table {%
0 88.4971923828125
0.0500000007450581 87.6693572998047
0.100000001490116 87.015998840332
0.150000005960464 86.5164642333984
0.200000002980232 86.0372619628906
0.25 85.5976638793945
0.300000011920929 85.0863494873047
0.349999994039536 84.4656219482422
0.400000005960464 83.8112106323242
0.449999988079071 83.0957794189453
0.5 82.2451324462891
0.550000011920929 81.1807632446289
0.600000023841858 79.6946334838867
0.649999976158142 77.4884338378906
0.699999988079071 74.1404037475586
0.75 69.3446502685547
0.800000011920929 61.7947883605957
0.850000023841858 50.2786407470703
0.899999976158142 31.9030914306641
0.949999988079071 7.54735851287842
1 0
};
\addlegendentry{MixFormer-\algNameShort [70.2]}
\addplot [ colorMT, densely dashed]
table {%
0 87.554801940918
0.0500000007450581 86.7252426147461
0.100000001490116 86.2041702270508
0.150000005960464 85.7693099975586
0.200000002980232 85.3878479003906
0.25 84.9785537719727
0.300000011920929 84.4200286865234
0.349999994039536 83.7032928466797
0.400000005960464 83.0398941040039
0.449999988079071 82.180290222168
0.5 81.2691955566406
0.550000011920929 80.1758499145508
0.600000023841858 78.6797485351562
0.649999976158142 76.4732666015625
0.699999988079071 73.2877044677734
0.75 68.499397277832
0.800000011920929 61.2012023925781
0.850000023841858 49.7166595458984
0.899999976158142 31.4170246124268
0.949999988079071 7.57318353652954
1 0
};
\addlegendentry{MixFormer [69.4]}
\addplot [ colorM,densely dotted]
table {%
0 86.7928009033203
0.0500000007450581 85.800895690918
0.100000001490116 85.0984191894531
0.150000005960464 84.542236328125
0.200000002980232 84.0398254394531
0.25 83.5353164672852
0.300000011920929 82.9559860229492
0.349999994039536 82.2148590087891
0.400000005960464 81.5032806396484
0.449999988079071 80.6581420898438
0.5 79.6057205200195
0.550000011920929 78.3901214599609
0.600000023841858 76.753662109375
0.649999976158142 74.3666763305664
0.699999988079071 70.9551315307617
0.75 66.0943603515625
0.800000011920929 58.6971282958984
0.850000023841858 47.4491767883301
0.899999976158142 29.8106555938721
0.949999988079071 7.31007623672485
1 0
};
\addlegendentry{MixFormer(w/oTU) [67.9]}

\addplot [ colorSB]
table {%
0 87.483283996582
0.0500000007450581 86.7001647949219
0.100000001490116 86.1699295043945
0.150000005960464 85.6908798217773
0.200000002980232 85.1894760131836
0.25 84.6287841796875
0.300000011920929 84.0730209350586
0.349999994039536 83.4442825317383
0.400000005960464 82.615837097168
0.449999988079071 81.5559844970703
0.5 80.3347854614258
0.550000011920929 78.8397064208984
0.600000023841858 77.0867004394531
0.649999976158142 74.5261154174805
0.699999988079071 70.7793655395508
0.75 65.1598358154297
0.800000011920929 56.6021499633789
0.850000023841858 43.7479629516602
0.899999976158142 25.0748329162598
0.949999988079071 5.10387992858887
1 0
};
\addlegendentry{STARK-\algNameShort [67.8]}
\addplot [colorST, densely  dashed]
table {%
0 85.6929626464844
0.0500000007450581 84.6479644775391
0.100000001490116 84.0160675048828
0.150000005960464 83.502326965332
0.200000002980232 82.9854125976562
0.25 82.3585205078125
0.300000011920929 81.4644165039062
0.349999994039536 80.574089050293
0.400000005960464 79.6707534790039
0.449999988079071 78.8283615112305
0.5 77.7788467407227
0.550000011920929 76.5073318481445
0.600000023841858 74.7033386230469
0.649999976158142 72.1949234008789
0.699999988079071 68.7872924804688
0.75 63.6377334594727
0.800000011920929 55.5557060241699
0.850000023841858 43.4120063781738
0.899999976158142 25.3304920196533
0.949999988079071 5.57294273376465
1 0
};
\addlegendentry{STARK-ST [66.1]}

\addplot [colorS,densely dotted]
table {%
0 85.6302947998047
0.0500000007450581 84.4858245849609
0.100000001490116 83.7390441894531
0.150000005960464 83.1157760620117
0.200000002980232 82.5374298095703
0.25 81.879508972168
0.300000011920929 80.9960098266602
0.349999994039536 80.121452331543
0.400000005960464 79.1458740234375
0.449999988079071 78.2955856323242
0.5 77.2004699707031
0.550000011920929 75.8454895019531
0.600000023841858 73.9936447143555
0.649999976158142 71.3914260864258
0.699999988079071 67.9372406005859
0.75 62.804084777832
0.800000011920929 54.8048095703125
0.850000023841858 42.9958000183105
0.899999976158142 25.6755027770996
0.949999988079071 5.70087146759033
1 0
};
\addlegendentry{STARK-S [65.6]}
\end{axis}

\end{tikzpicture}}
   \resizebox{0.85\linewidth}{!}{\input{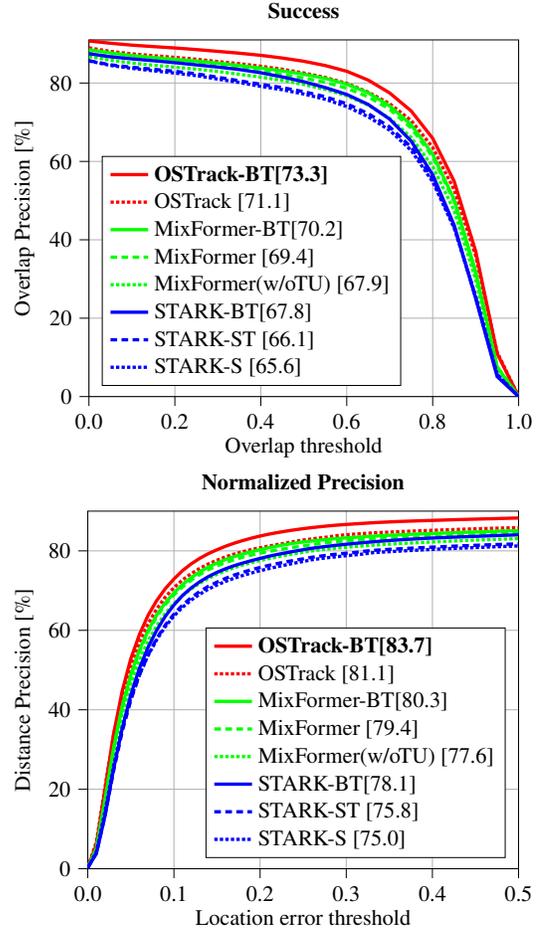}}
   \caption{Plots of success and normalized precision for the proposed methods and various SOTA trackers on LaSOT. The curves of the tracker with \algName~(lined) are above each curves from the base model without template update (dotted) and the base model with their own template update method (dashed). }
   \label{fig:graph_LaSOT}
\end{figure}

Fig.~\ref{fig:graph_LaSOT} shows the evaluation results of the performance of the proposed method, \algName. We applied the propewosed method to the existing SOTA trackers~\cite{ye2022joint,cui2022mixformer,yan2021learning} and conducted experiments on LaSOT~\cite{fan2019lasot}, a representative long-term tracking dataset. When the proposed method was applied to OSTrack~\cite{ye2022joint}, the score of success and normalized precision ($P_{Norm}$) increased 2.2\% and 2.7\%, respectively. On the other hand, MixFormer~\cite{cui2022mixformer} and STARK~\cite{yan2021learning} have their own template update method. 
In the case of MixFormer, when our method (MixFormer-\algNameShort) replaced with existing template update (MixFormer), success and $P_{Norm}$ improved 0.8\% and 0.9\%, respectively. And compared to the base tracker without template update, MixFormer(w/o TU) in Fig.~\ref{fig:graph_LaSOT}, success and $P_{Norm}$ improved 2.3\% and 2.7\%, respectively. Similarly, STARK-\algNameShort~improved 1.7\% in success and 2.3\% in $P_{Norm}$ when replacing the existing template update (STARK-ST) with \algName, and improved 2.2\% in success and 3.1\% in $P_{Norm}$ compared to the base tracker (STARK-S).

\subsection{\algName~on VOT2022 challenge}
\subsubsection{Details of the tracker} 
We modified STARK~\cite{yan2021learning} as the base model for this dataset. The backbone network is replaced by Swin Transformer-Large~\cite{liu2021swin} and we applied \algName~for robust and precise template update. \algName~try to update the template more frequently ($N$=7) compared to the original STARK-ST~\cite{yan2021learning}. Additionally, Alpha-Refine~\cite{yan2021alpha} is applied for each of the bounding box output to improve the box estimation quality of the tracker. 

This tracker with \algName~scores on par with the best-ranking in \emph{Unsupervised} reference scores at the short-term bounding box regression track of VOT2022 challenge~\cite{Kristan2022vot}.
\emph{Unsupervised} scores is a evaluation of tracker without \emph{short-term failure recovery}.

\subsubsection{\emph{Unsupervised} reference scores}
For the short-term track of VOT2022~\cite{Kristan2022vot}, the evaluation criterion focuses more on how accurately the tracker locates the bounding box well. Therefore, if a tracker miss the target several frames, the tracker is re-initialize with the ground truth information and restarts the tracking with new initialized position. 

It means that the \emph{short-term failure recovery} not only accesses the bounding box ground truth for re-initialization but also continuously accesses the ground truth bounding box for every frame to check whether the tracker misses the target or not. This is an unusual evaluation setting for real world application because we cannot access the ground truth information for the real world uses. 

Therefore, \emph{Unsupervised} scores which were evaluated without the \emph{short-term failure recovery} is more usual evaluation setting for trackers and same as the evaluation protocol of the other object tracking benchmarks~\cite{fan2019lasot,muller2018trackingnet,huang2019got}. And our tracker with \algName~scores on par with the best-ranking in those \emph{Unsupervised} reference scores.

The results of VOT-STb2022 track are shown in Table.~\ref{table:VOTchallenge}. Please, refer~\cite{Kristan2022vot} for more details of the other trackers.

\vspace{-0.2cm}
\begin{table}[h!]
\centering
\caption{Results of VOT2022 for Short-Term bounding box track 
}
\begin{adjustbox}{width=0.90\linewidth,center}

\begin{tabular}{c|c|ccc}
\hline
VOT-STb2022 & \ti{Unsupervised} & \multicolumn{3}{c}{\ti{Baseline}}                                                            \\ \hline
Tracker                               & AUC          & \multicolumn{1}{c}{EAO} & \multicolumn{1}{c}{accuracy} & \multicolumn{1}{c}{robustness} \\ \hline
Proposed                              & \textbf{0.735}                   & 0.569                   & 0.775                        & 0.862                          \\
APMT\_RT                              & 0.721                            & 0.581                   & 0.787                        & 0.877                          \\
DAMT                                  & 0.716                            & 0.602                   & 0.776                        & \textbf{0.887}                 \\
MixFormerL                            & 0.708                            & \textbf{0.602}          & \textbf{0.831}               & 0.859                          \\
OSTrackSTB                            & 0.680                            & 0.591                   & 0.790                        & 0.869                          \\ \hline
\end{tabular}
\end{adjustbox}
\label{table:VOTchallenge}
\end{table}
\vspace{-0.4cm}

\subsection{Hit ratio of Backward tracking}
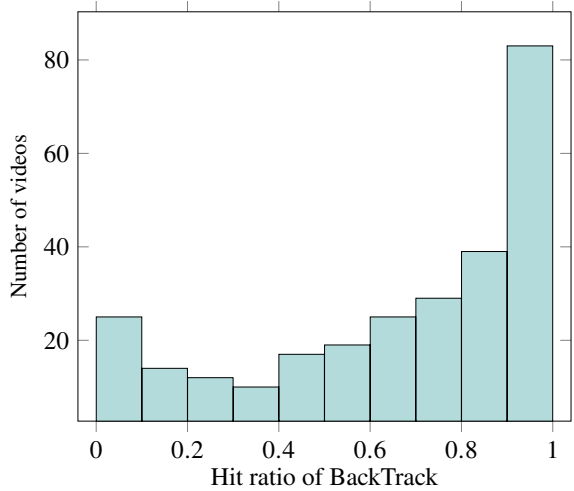
\begin{figure}[!h]
   \centering
   \begin{adjustbox}{width=0.9\columnwidth,center}
\begin{tikzpicture}
\begin{axis} [ybar,
    bar width=18pt,
    ylabel={Number of videos}, 
    ylabel near ticks,
    ylabel style={font=\small},
    xtick style={font=\Huge},
    xlabel={Hit ratio of \algName}
]
\addplot[fill=colorbase] coordinates {
    (0.05,  25) 
    (0.15,  14) 
    (0.25,  12)
    (0.35,  10)
    (0.45,  17)
    (0.55,  19)
    (0.65,  25)
    (0.75,  29)
    (0.85,  39)
    (0.95,  83)
};
\end{axis}
\end{tikzpicture}\end{adjustbox}
   \caption{Histogram for hit ratio of OSTrack-\algNameShort~on LaSOT. The test set of LaSOT has a total of 280 videos.}
   \label{fig:histo}
\end{figure}

The proposed method shows 65.5\% of average hit ratio on LaSOT ($N$=15, $\iouzerothres$=0.9) which means \algName~tries a hundred times of template updates and accepts the candidate 65 times (i.e. rejects 35 times) in average. Fig.~\ref{fig:histo} shows the hit ratio of OSTrack\algNameShort~on LaSOT. \algName~continuously updated its template in more than the 80 sequences of LaSOT and about the 20 videos were updated less than 10\% of their tries of template updates.  

\subsection{Hyper-parameters of \algName}
There are two hyper-parameters for \algName($N$,$\iouzerothres$); $N$ is the number of frame to performs backward tracking and $\iouzerothres$ is the threshold of IOU between the first forward frame and last backward frame.
OSTrack-\algNameShort~shows the best performance with (15,0.9). $\iouzerothres$=0.9 means the template is updated only if the two boxes overlap almost exactly. STARK-\algNameShort~shows the best performances with (20,0.9) on LaSOT and GOT-10k~\cite{huang2019got}, (10,0.9) for $\text{LaSOT}_{ext}$~\cite{fan2021lasot} and (20,0.7) for TrackingNet~\cite{muller2018trackingnet}.

\begin{table}[h!]
\centering
\caption{Performance improvements (AUC \%) of OSTrack using \algName~with the combination of hyper-parameters ($N$ and $\iouzerothres$). The AUC of the baseline is 71.1\%. }
\begin{adjustbox}{width=\columnwidth}
\begin{tabular}{c|rrrrrrrrr} 
\hline
\multicolumn{1}{l|}{\backslashbox{ $\iouzerothres$ }{$N$}} & 5 & 10 & 15 & 20 & 25 & 30 \\ 
\hline
0.7 & 0.0 & 1.0 & 0.7 & 1.1 & 0.5 & 0.8\\
0.8 & 1.0 & 1.9 & 2.0 & 1.8 & 1.6 & 1.9\\
0.9 & 1.0 & 1.7 & \tb{2.2} & 1.5 & 1.4 & 1.7\\
\hline
\end{tabular}
\end{adjustbox}
\label{table:HPsearch}
\end{table}
As shown in Table.~\ref{table:HPsearch}, \algName~consistently improves the performance of the base tracker regardless of hyper-parameters, $N$ or $\iouzerothres$. \algName~has some variations on scores of improvement from 0.5\% to 2.2\%. However, the proposed method shows robustness independent of hyper-parameters, while the performances of the other template update methods~\cite{yan2021learning,cui2022mixformer} were degraded according to the choice of the hyper-parameter $N$.

\subsection{Meaning of $N$ and $\sigma^{thres}$}
As shown in Fig.1 of the main paper, more frequent updates boosts more performance of the tracker only if there is less false positive case. However, \algName~shows less improvement when it uses too short updates cycle such as $N$=5 compared to $N$=10 or $N$=15 cases because \algName~is getting more robust and accurate when it has enough length of tracklets to compare each other. Therefore, the trade off between the improvement by the frequent updates and robustness of the template update decision.





\subsection{Inference speed of \algName}
\begin{table}[h!]
\centering
\begin{adjustbox}{width=0.9\linewidth,center}

\begin{tabular}{l|r|cc} 
\hline
Model & MACs & FPS & FPS* \\ 
\hline
Base model   &  60.6 G  & 64.9 & 94.1 \\ 
+\algName384 & 121.2 G  & 29.0 & 47.0 \\
+\algName128 &  67.4 G   & 35.4 & 66.6 \\ 
+\algName64  &  62.5 G   & 36.2 & 69.4 \\
\hline
\end{tabular} 
\end{adjustbox}
\caption{ Average computational costs per frame (MACs) and the inference speeds (FPS \& FPS*) of the trackers on LaSOT without the early termination and the early rejection. The base model is OSTrack384. FPS (frames per second) is the inference speed of the tracker including pre-processing time and FPS* represents the speed for only model inference.}
\label{table:supple_FPS}
\end{table}

MACs of the backward tracker (\algName64 \& \algName128) significantly were reduced to 3\% \& 11\% of \algName384, respectively. However, the average FPS between the base model and \algName64~show a big difference even if their MACs are similar as shown in Table.\ref{table:supple_FPS}.
This is because we compute the inference speed of \algName~from the inference time, including a series of input image crops and pre-possessing for Backward tracking. (FPS in Table.~\ref{table:supple_FPS}). Calculating the speed excluding the time required for data crop and pre-processing (FPS* in Table.~\ref{table:supple_FPS}) is as follows; 47.0 for \algName384, 66.6 for \algName128, and 69.4 for \algName64. In the main paper, for the fair comparison, we published the average FPS calculated from the time which includes whole time for image crops and pre-processing.

\subsection{Combination of \algName~and the original confidence head}
We examined the possibility of combination of \algName~and the original template update method (confidence head) of MixFormer. When both the original template update rule of MixFormer and \algName~are applied simultaneously, there was 0.2\% of AUC gain on LaSOT compared to \algName~only case. However, on LaSOT$_{ext}$, AUC decreased by 1.0\% under the combined update rule compared to the AUC of \algName~only. We conclude that there is no feasible synergy between the original confidence head and \algName~because the original MixFormer does not provide robust and effective template update rule.

\subsection{Hybrid Backward tracking with non-deep learning based tracker}

\begin{table}[h!]
\centering
\begin{adjustbox}{width=0.9\linewidth}
\begin{tabular}{lcll}
\hline
\multirow{2}{*}{Method} & \multirow{2}{*}{\begin{tabular}[c]{@{}c@{}}Template\\ update\end{tabular}}  & \multicolumn{2}{c}{LaSOT}     \\ \cline{3-4} 
                        &                                                                            & AUC            & $P_{Norm}$    \\ \hline
OSTrack           & No       & 70.30             & 80.62          \\
OSTrack-BT(KCF)   & Proposed & 71.32\myUpR{1.02} & 81.31\myUpR{0.69} \\ \hline
\end{tabular}
\end{adjustbox}
\vspace{-0.2cm}
\caption{Performance improvement by BackTrack with KCF on LaSOT. The numbers between parenthesis represent the performance improvement over each base tracker. 
}
\label{table:KCF}
\end{table}

\algName~is generally applicable on any type of tracker. Since the quantitative metric of the proposed method is based on the IOU comparison, any type of tracker can be used as a backward tracker. Kernelized Correlation Filter (KCF)~\cite{henriques2014high} is not a deep learning based tracker.  Here, we applied KCF as backward tracker on the deep learning based tracker (OSTrack). The performance of KCF is far from the SOTA performance especially on the the standard benchmarks for recent VOT studies such as LaSOT, TrackingNet, etc. However, OSTrack-BT(KCF) improves the performance 1.02 \% in AUC and 0.69 \% in norm precision as shown in Table.\ref{table:KCF}. While KCF shows inferior performance compared to the recent deep trackers, the template update by the proposed method robustly improves the forward tracker with the proposed algorithm.


\twocolumn[{%
\renewcommand\twocolumn[1][]{#1}%

\subsection{Python code}
Here, we present the code of BackTrack for the reproducibility of the proposed method. BackTrack is able to be implemented with the simple modification of the original deep tracker as followings.
\lstset{language=Python}
\lstset{frame=lines}
\lstset{caption={Backward Track source code. The proposed algorithm can be easily added to existing deep trackers by simply editing the code. The definitions of some functions(`MyTracker', `Crop', etc) are omitted for the simplicity of the code.}}
\lstset{label={lst:code_direct}}
\lstset{basicstyle=\footnotesize}
\begin{lstlisting}
from math import floor
class MyTracker_with_BackTrack:
    # X: RGB image, B: bounding box
    def __init__(self):
        self.model       = MyTracker() # Tracker 
        self.N           = 15          # Period of checking template update
        self.sigma_thres = 0.9         # Threshold for Backward tracking metric        
        
    def initialize(self, X0, B0):
        Z0             = Crop(X0, B0)  # Crop template Z0 from the image X using the bounding box B
        self.Z         = [Z0, Z0]      # Dual templates. Simplely initialized as the same template Z0
        self.frame_idx = 0             # Index of current frame
        self.k_step    = 1             # Sampling step for Backward tracking
        self.t_start   = 0             # Backward tracking runs until it reaches t_start
        self.M_thres   = floor(self.N*self.sigma_thres)
        self.min_sz_Z  = 64**2         # Minimum size(area) of template for early rejection
        self.X_list    = [X0]          # Store images for Backward tracking as list
        self.B_list    = [B0]          # Store bounding boxes for Backward tracking as list
        
    def track(self, X):
        self.frame_idx +=1
        # Track the target in current image X using the templates stored
        B_pred = self.model(X, self.Z) 
        self.X_list.append(X)
        self.B_list.append(B_pred)
        # Backward tracking goes here 
        early_reject = (B_pred.width * B_pred.height < self.min_sz_Z) # template is too small
        if (self.frame_idx % self.N) == 0 and (not tag_early_reject): 
            # Extracted from current X using the predicted bounding box
            Z_candi   = Crop(X, B_pred) 
            M, Sigma0 = self.backtrack(Z_candi)
            if (M > self.M_thres) and (Sigma0 > self.sigma_thres): 
                # Do template update  
                self.Z[1]    = Z_candi           
                self.t_start = self.frame_idx
                self.k_step  = 1
            else: 
                # Decide not to update the template, increase the sampling rate
                self.k_step+=1
        return B_pred
        
    def backtrack(self, Z_candi):
        # Sampled the list of X and B with a rate of k_step from t_start to current 
        X_list = self.X_list[self.t_start:self.k_step:]
        B_list = self.B_list[self.t_start:self.k_step:]   

        M = 0 # Number of frames which has enough overlap between fwd & bwd track
        t = len(X_list)
        while t>=0:
            t-=1    # Negative means backward
            # Backward tracking using only candidate template without Z0 template 
            B_bwd = self.model(X_list[t], Z_candi) 
            if IOU(B_bwd, B_list[t])>0.5:
                M+=1    # Enough IOU between fwd & bwd track
            else:
                return M, 0.    # Early Temination 
        Sigma0 = IOU(B_bwd, B_list[0])  # Here, IOU of t_start-th frame
        return M, Sigma0
        
\end{lstlisting}

}]

\twocolumn[{%
\renewcommand\twocolumn[1][]{#1}%

\lstset{language=Python}
\lstset{frame=lines}
\lstset{caption={Main code for running the tracker with BackTrack.}}
\lstset{label={lst:code_direct}}
\lstset{basicstyle=\footnotesize}
\begin{lstlisting}
def main():
    MyTrackBT = MyTracker_with_BackTrack()
    # X0    : RGB image of first frame for initialization
    # B0    : Bounding box of first frame for initialization
    # X_list: Sequence of RGB images for tracking
    MyTrackBT.initialize(X0,B0)
    B_list = [] # prediction of bounding boxes
    for X_t in X_list:
        B_pred = MyTrackBT.track(X_t)
        B_list.append(B_pred)
    return B_list
\end{lstlisting}

}]



\end{document}